\documentclass{article}

\usepackage{arxiv}

\usepackage[utf8]{inputenc} 
\usepackage[T1]{fontenc}    
\usepackage{hyperref}       
\usepackage{url}            
\usepackage{booktabs}       
\usepackage{amsfonts}       
\usepackage{amsmath} 
\usepackage{amssymb}  
\usepackage{nicefrac}       
\usepackage{microtype}      
\usepackage{cleveref}       
\usepackage{lipsum}         
\usepackage{graphicx}
\usepackage{natbib}
\usepackage{doi}
\usepackage{epsfig} 
\usepackage{mathptmx} 
\usepackage{times} 
\usepackage{bm}

\usepackage{subfig}
\usepackage{multirow}
\usepackage[export]{adjustbox}
\usepackage{url}
\usepackage[usenames]{color}
\usepackage[mathscr]{euscript}

\title{Scaled 360 layouts: \\Revisiting non-central panoramas}

\date{}

\author{ \href{https://orcid.org/0000-0003-2674-4844}{\includegraphics[scale=0.06]{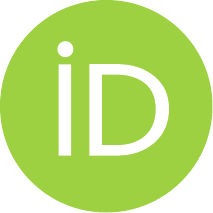}\hspace{1mm}Bruno Berenguel-Baeta}\thanks{Corresponding author.} \\
	Instituto de Investigacion en Ingenieria de Aragon\\
	Department of Computer Science and Systems Engineering\\
	University of Zaragoza\\
	Zaragoza, Spain \\
	\texttt{berenguel@unizar.es} \\
	\And
	\href{https://orcid.org/0000-0002-8479-1748}{\includegraphics[scale=0.06]{orcid.pdf}\hspace{1mm}Jesus Bermudez-Cameo} \\
	Instituto de Investigacion en Ingenieria de Aragon\\
	Department of Computer Science and Systems Engineering\\
	University of Zaragoza\\
	Zaragoza, Spain \\
	\texttt{bermudez@unizar.es} \\
	\And
	\href{https://orcid.org/0000-0001-5209-2267}{\includegraphics[scale=0.06]{orcid.pdf}\hspace{1mm}Jose J. Guerrero} \\
	Instituto de Investigacion en Ingenieria de Aragon\\
	Department of Computer Science and Systems Engineering\\
	University of Zaragoza\\
	Zaragoza, Spain \\
	\texttt{josechu.guerrero@unizar.es} \\
}
\renewcommand{\shorttitle}{\textit{Scaled 360 layouts}}
\newcommand\blfootnote[1]{%
  \begingroup
  \renewcommand\thefootnote{}\footnote{#1}%
  \addtocounter{footnote}{-1}%
  \endgroup
}
\hypersetup{
pdftitle={Scaled 360 layouts: Revisiting non-central panoramas},
pdfsubject={cs.CV},
pdfauthor={Bruno Berenguel-Baeta, Jesus Bermudez-Cameo, Jose J. Guerrero},
pdfkeywords={Omnidirectional Vision; 3D Vision; Non-central Cameras; Layout recovery; Scene understanding},
}

\begin{document}
\maketitle
\blfootnote{A final version of this article can be found at \url{https://doi.org/10.1109/CVPRW53098.2021.00410}}

\begin{abstract}
From a non-central panorama, 3D lines can be recovered by geometric reasoning. However, their sensitivity to noise and the complex geometric modeling required has led these panoramas being very little investigated. In this work we present a novel approach for 3D layout recovery of indoor environments using single non-central panoramas.  We obtain the boundaries of the structural lines of the room from a non-central panorama using deep learning and exploit the properties of non-central projection systems in a new geometrical processing to recover the scaled layout. We solve the problem for Manhattan environments, handling occlusions, and also for Atlanta environments in an unified method. The experiments performed improve the state-of-the-art methods for 3D layout recovery from a single panorama. 
Our approach is the first work using deep learning with non-central panoramas  and recovering the scale of single panorama layouts.
\end{abstract}


\section{Introduction}

Layout recovery and 3D understanding of indoor environments is a hot topic in computer vision research \cite{zou2021manhattan}. Most recent approaches for layout recovery use different neural network architectures to recover the structural elements of an indoor environment. In this context, the use of omnidirectional images has important advantages to retrieve the shape of a whole room from a single image. In order to handle the heavy distortions that introduce the omnidirectional representations, we can find different approaches in the state of the art. Dula-Net \cite{yang2019dula} extracts from the equirectangular image a perspective view of the ceiling and extrude the layout using a dual-branched architecture. Similarly but with a different network architecture, AtlantaNet \cite{pintore2020atlantanet} obtains the floor plan dividing the panorama into two perspective views of ceiling and floor separately to adjust the height of the room. On a different approach, Corners for Layouts (CFL) \cite{fernandez2020corners} and HorizonNet \cite{sun2019horizonnet} aim to extract the boundaries of the structural lines and corners of the room directly from the equirectangular panorama, obtaining an up-to-scale layout in a post-processing. 

\begin{figure}
	\centering
	\includegraphics[width=0.47\textwidth]{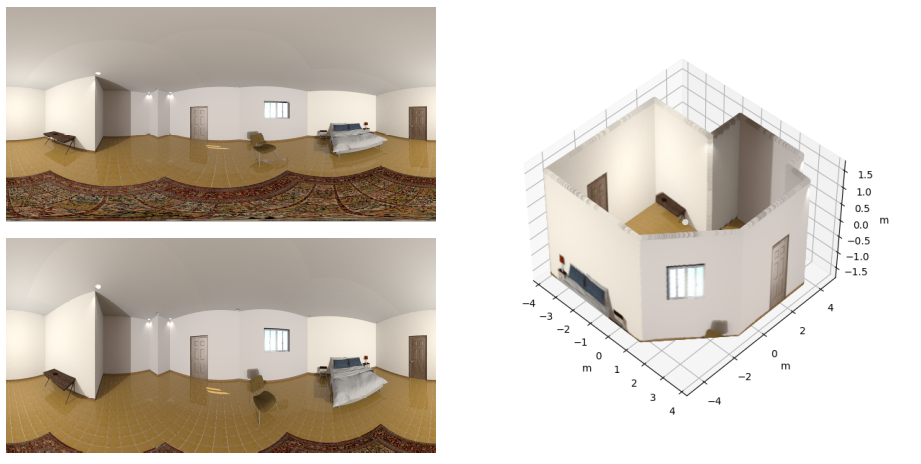}
	\caption{Central (up-left) and non-central panoramas (bottom-left) have similar appearance but there are subtle differences in favor of the second if we want to obtain 3D information including the scale. On the right, the scaled layout obtained from a non-central panorama in an Atlanta environment with our solution.}
	\label{fig:intro}
\end{figure}

In this paper we propose a new method for layout recovery from single panoramas. We revisit the non-central panoramas \cite{li2004stereo} in order to obtain scaled 3D lines \cite{bermudez2017exploiting} and vertical planes from a single image. As the equirectangular panorama, the non-central circular panorama provides 360 information of the environment, but the image distortion in non-central panoramas includes subtle differences allowing geometric 3D reasoning \cite{perdigoto2012reconstruction}. This characteristic is a clear advantage with regard equirectangular panoramas since allows to recover the scale of the environment without any prior knowledge. 
In Fig. \ref{fig:intro} we have an equirectangular panorama and a non-central panorama of the same scene. Even though both images look similar allowing to recover the shape of the room, only from the non-central circular panorama we can recover the scale without any assumption. 

\begin{figure*}
	\centering
	\includegraphics[width = \textwidth]{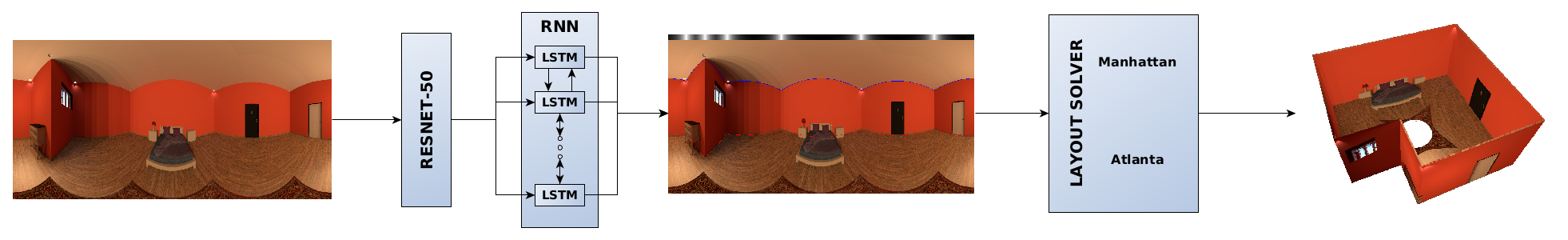}
	\caption{Pipeline of the proposed method. The non-central circular panorama is processed by the fine-tuned network. The network provides the pixel information of the structural lines and a per-column probability of a wall-wall intersection. Then the proposed geometric pipeline, including the new solvers, gives the final scaled layout.}
	\label{fig:pipeline}
\end{figure*}

In our proposal, we adapt the neural network architecture of HorizonNet \cite{sun2019horizonnet} to non-central circular panoramas for the extraction of structural lines of indoor environments. Besides, we propose a geometric pipeline with two new linear solvers that jointly obtain the room height and vertical walls location for global layout extraction in Manhattan and Atlanta world assumptions. Our experiments show that our proposal improves the state of the art solutions, being ours the first that obtains the scale of the layout without using additional assumptions.


\section{Proposed method}
\label{sec:method}

In order to recover the layout of a room from a single non-central circular panorama, we propose a new pipeline, shown in Fig. \ref{fig:pipeline}. We use a neural network as a line extractor from the image to reconstruct the scaled layout in a new geometrical processing that includes two new linear solvers for non-central projection systems.

\subsection{Neural network as line extractor}
\label{subsec:network}

For the first part of our pipeline, we propose a neural network as boundary extractor for structural lines. In classical approaches, lines are extracted with Hough transform and vanishing points, followed by an hypothesis generation-verification algorithm \cite{zhang2014panocontext}. This approach consumes lots of time and resources. On the other hand, neural networks have proven that can obtain patterns on images with high accuracy in a short time and, therefore, current approaches for layout recovery rely on the use of neural networks.

Even though many state-of-the-art networks can handle omnidirectional panoramas, there are not any that have considered non-central systems. We propose to adapt the existing network architecture of HorizonNet \cite{sun2019horizonnet} to handle non-central panoramas. The main advantage of this architecture is that handles the information of the panorama column by column. This is particularly interesting for the non-central circular panorama since it is locally a central projection system in each column. 

In order to adapt the network to the distortions of the non-central panoramas, we have fine-tuned it. However, since non-central projection systems are little used in the research community, there is no data-set available. To overcome this difficulty, we have generated a data-set of non-central circular panoramas with ground truth information for layout recovery in synthetic indoor environments. This data-set is composed of around 650 different layouts from 6 to 10 walls and more than 2500 images. The data-set will be available under request.

\subsection{Geometric solvers}
\label{subsec:geom}

In the second part of our pipeline, we take the pixel information provided by the network and reconstruct the scaled layout of the room. For that purpose, we derive two new geometric solvers to jointly obtain the room height and vertical walls location for Manhattan or Atlanta world assumptions.

\begin{figure}
	\centering
	\includegraphics[width = 0.3\textwidth]{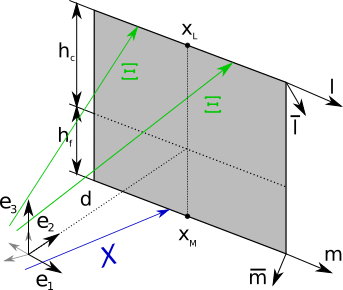}
	\caption{Rays and wall parameter definition. The parameters are: wall reference system $\{\mathbf{e}_1,\mathbf{e}_2,\mathbf{e}_3\}$; $\boldsymbol{\Xi}$ and $\boldsymbol{X}$ define the projecting rays; $\mathbf{(l,\bar{l})}$ and $\mathbf{(m,\bar{m})}$ are the ceiling and floor lines that define the wall; $\mathbf{x_L,x_M}$ define the closest points of the lines to the origin; $h_c$, $h_f$ and $d$ are the ceiling and floor height and distance to the wall respectively.}
	\label{fig:atlantasolver}
\end{figure}

We define a wall as a set of two parallel lines contained in a vertical plane (see Fig. \ref{fig:atlantasolver}). Let $\mathbf{L} = (\mathbf{l}^T,\mathbf{\bar{l}}^T)^T$ and $\mathbf{M} = (\mathbf{m}^T,\mathbf{\bar{m}}^T)^T$ be the ceiling and floor lines defined in Plücker coordinates \cite{pottmann2009computational} and $\{\mathbf{e}_1,\mathbf{e}_2,\mathbf{e}_3\}$ an orthonormal basis attached to the vertical wall. We define the closest points of the lines to the acquisition system $\mathbf{x}_L$ and $\mathbf{x}_M$ with $h_c$ and $h_f$, distance from the acquisition system to the ceiling and floor planes, and $d$, distance to the wall plane, such that $\mathbf{x}_L = d \mathbf{e}_2 + h_c \mathbf{e}_3$ and $\mathbf{x}_M = d \mathbf{e}_2 + h_f \mathbf{e}_3$. Notice that with this description we can parameterize the Plücker coordinates of the lines as $\mathbf{l} = \mathbf{m} = \mathbf{e}_1$, $\mathbf{\bar{l}} = \mathbf{x_L} \times \mathbf{l} = h_c \mathbf{e}_2 - d \mathbf{e}_3$, 
$\mathbf{\bar{m}} = \mathbf{x_M} \times \mathbf{m} = h_f \mathbf{e}_2 - d \mathbf{e}_3$.  
We also define the projecting rays that intersect the ceiling and floor lines as $\boldsymbol{\Xi} = (\boldsymbol{\xi}^T,\boldsymbol{\bar{\xi}}^T)^T$ and $\boldsymbol{X} = (\boldsymbol{\chi}^T,\boldsymbol{\bar{\chi}}^T)^T$ respectively. 

\begin{equation}\label{eq:side_l_xi}
	side(\boldsymbol{\Xi},\mathbf{L}) = \boldsymbol{\xi}^T \left( h_c \mathbf{e}_2 - d \mathbf{e}_3 \right) + \boldsymbol{\bar{\xi}}^T \mathbf{e}_1 = 0
\end{equation}
\begin{equation}\label{eq:side_m_chi}
	side(\boldsymbol{X},\mathbf{M}) = \boldsymbol{\chi}^T \left( h_f \mathbf{e}_2 - d \mathbf{e}_3 \right) + \boldsymbol{\bar{\chi}}^T \mathbf{e}_1 = 0
\end{equation}

Known the projecting rays, given by the output of the neural network, we aim to obtain the 3D lines that define each wall in the environment. The relation among the projection rays and the lines is given by their intersection, defined in equations \eqref{eq:side_l_xi} and \eqref{eq:side_m_chi}. 
This is, in general, a non-linear problem which is difficult to tackle directly. However, we propose two new DLT-like approaches that allows to compute the solution for the layout as a linear problem.

\begin{equation}\label{eq:manx}
	\bar{\xi}_{1}  u_x + \bar{\xi}_{2}  u_y - \xi_{1}  v_y - \xi_{2}  v_x  - d \xi_{3}  = 0
\end{equation}
\begin{equation}\label{eq:many}
	\bar{\chi}_{1}  u_x - \bar{\chi}_{2}  u_y - \chi_{1}  w_y - \chi_{2}  w_x  - d \chi_{3}  = 0
\end{equation}

Let the main direction of a wall be horizontal and described by the vector $\mathbf{u} = \left(u_x,u_y\right)^T$ such that $\mathbf{l} = \mathbf{m} = (u_x,u_y,0)^T$. We can define vectors $\mathbf{v} = h_c \mathbf{u}$ and $\mathbf{w} = h_f \mathbf{u}$ such that expressions \eqref{eq:side_l_xi} and \eqref{eq:side_m_chi} become linear obtaining a set of expressions (\eqref{eq:manx} and \eqref{eq:many}) depending on the unknown wall homogeneous vector $\mathbf{W} = \left( \mathbf{u}^T, \mathbf{v}^T, \mathbf{w}^T, d \right)^T$.

\begin{equation}\label{eq:lambda_uv}
 \lambda(\mathbf{v_1} - h_c \mathbf{u_1}) = h_c \mathbf{u_0} - \mathbf{v_0}
\end{equation}
\begin{equation}\label{eq:lambda_uw}
	 \lambda(\mathbf{w_1} - h_f \mathbf{u_1}) = h_f \mathbf{u_0} - \mathbf{w_0}
\end{equation}

However in this linear system $A \mathbf{W} = 0$, $\mathbf{u}$, $\mathbf{v}$ and $\mathbf{w}$ are independent variables which are non-parallel. In order to impose the parallelism of these vectors we compute the null space of the system with a Singular Value Decomposition (SVD) obtaining a parametric solution which is the linear combination of the singular vectors with a set of parameters $\lambda_i$. Two horizontal lines contained in a vertical plane have 4 degrees of freedom. 
A minimal solution would need 2 rays for each line of the wall, describing the null-space with three singular vectors and two parameters $\lambda_1$ and $\lambda_2$. By solving a system of two quadratic equations for $\lambda_1$ and $\lambda_2$ (with resultants, action matrices or as a polynomial eigenvalue problem \cite{kukelova2011polynomial}) we obtain a set of 4 different solutions which should be discriminated.
Since the network provides enough robust information, instead of the minimal solution, we propose to solve the over-determined case (with a minimum of 3 rays lying to each line) with a linear combination involving two singular vectors and single parameter $\lambda$ (such that $\mathbf{W}=\mathbf{W}_0+\lambda \mathbf{W}_1$) obtaining two uncoupled quadratic equations \eqref{eq:lambda_uv} and \eqref{eq:lambda_uw} respectively. These equations provide two solutions, where only one of them sets the ceiling line above the floor line.

Notice that with Manhattan world assumption there is a set of walls sharing the wall direction $\mathbf{u} = \left(u_x,u_y\right)^T$ and the complementary set of walls share the orthogonal direction vector $\mathbf{u}_\perp =\left(-u_y,u_x\right)^T$.
Since all the walls share the ceiling height $h_c$ and the floor $h_f$, we extend the DLT-like fitting to the whole set of walls, by computing the null-space of $\mathsf{A} \mathfrak{L}_M = 0 $ where $\mathfrak{L}_M$ is the layout vector $ \mathfrak{L}_M = \left( \mathbf{u}^T,\mathbf{v}^T, \mathbf{w}^T, d_1,\dotsm,d_N \right)^T $ where $N$ is the number of walls and the matrix $\mathsf{A}$ is full-filed with relations \eqref{eq:manx} and \eqref{eq:many}. The same reasoning as in the case of a single wall can be used to enforce parallelism among $\mathbf{u}$, $\mathbf{v}$, $\mathbf{w}$ .

\begin{equation}\label{eq:genx}
	\bar{\xi_1}' + h_c \xi_2' - d \xi_3' = 0
\end{equation}
\begin{equation}\label{eq:geny}
	\bar{\chi_1}' + h_f \chi_2' - d \chi_3' = 0
\end{equation}

For Atlanta world assumption each wall can have a different horizontal direction, therefore we have to extract each wall independently. Notice that in this case we are not imposing that $h_c$ and $h_f$ are common to all the walls. However, if the direction of each wall is known (for example extracting each wall independently) we can derive a new solution for the whole layout.
Assuming that wall directions are known, we can express the projecting rays of each wall in its own local reference system and then equations \eqref{eq:side_l_xi} and \eqref{eq:side_m_chi} become \eqref{eq:genx} and \eqref{eq:geny} respectively, where $\boldsymbol{\Xi}'$ and $\boldsymbol{X}'$ are the projecting rays in each wall reference system. 
Then, we can solve the null-space of a system of linear equations $\mathsf{A} \mathfrak{L}_A = \mathbf{0}$ with $\mathfrak{L}_A = \left(1,h_c,h_f,d_1,\dotsm,d_N\right)$ where $\mathsf{A}$ is composed from equations \eqref{eq:genx} and \eqref{eq:geny}.
The main advantage of this second approach is that can be used for Manhattan as well as Atlanta world environments whenever the layout has only one ceiling and floor heights.

\section{Experiments}

\begin{figure*}
	\centering
	\subfloat{\includegraphics[width=0.23\textwidth ,valign=c]{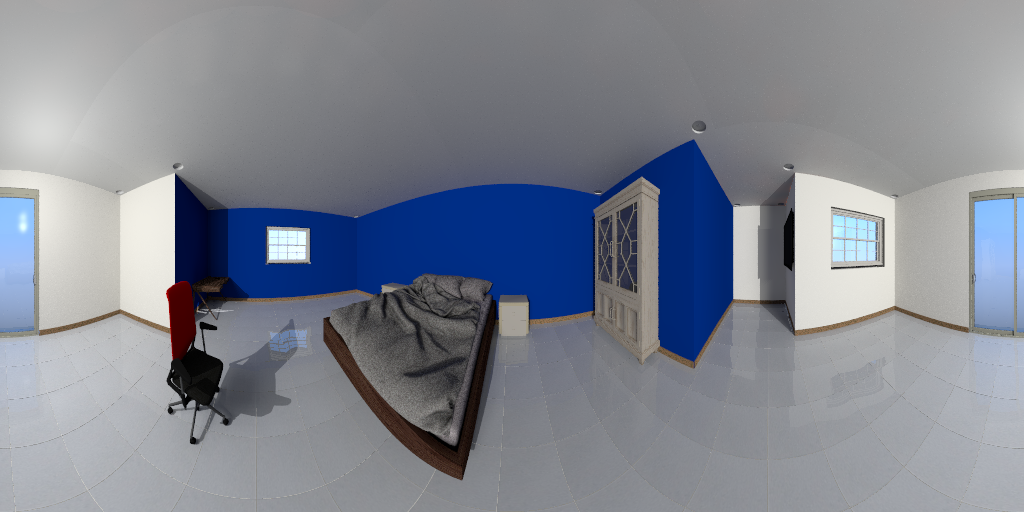}} \hfil
	\subfloat{\includegraphics[width=0.23\textwidth ,valign=c]{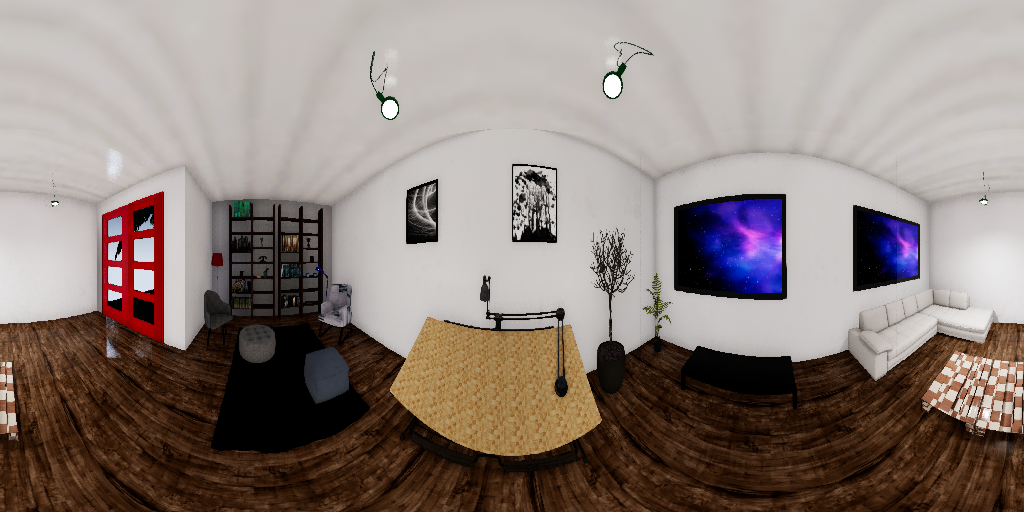}} \hfil
	\subfloat{\includegraphics[width=0.23\textwidth ,valign=c]{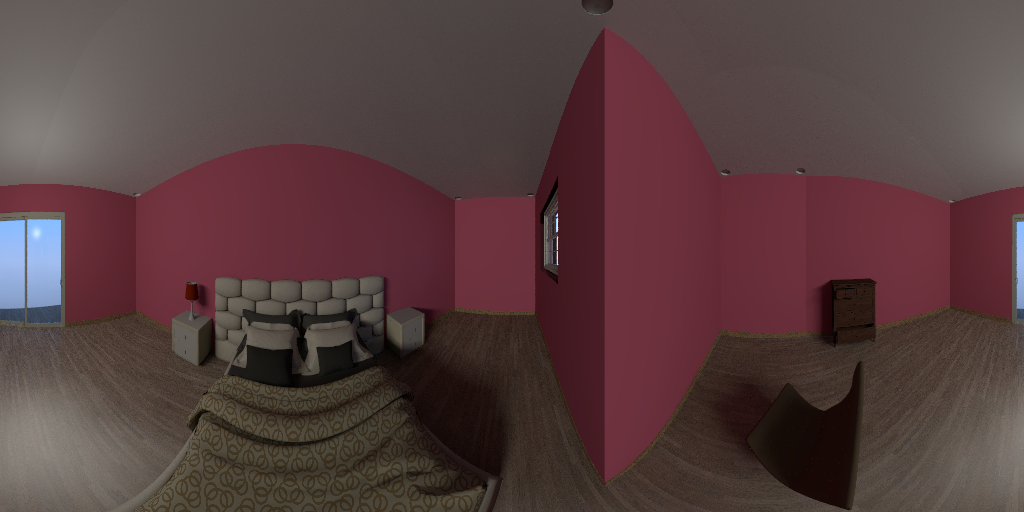}} \hfil
	\subfloat{\includegraphics[width=0.23\textwidth ,valign=c]{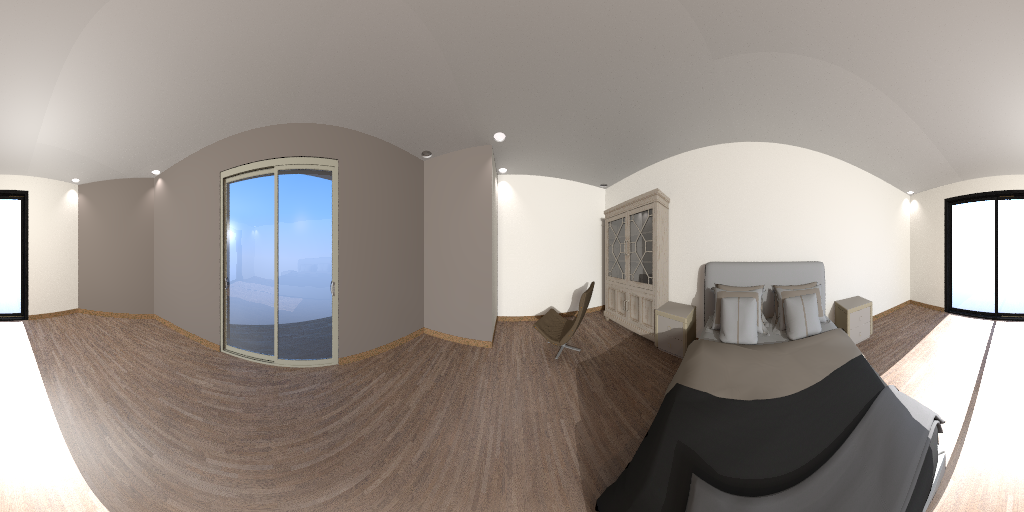}} \\
	
	\subfloat{\includegraphics[width=0.21\textwidth ,valign=c]{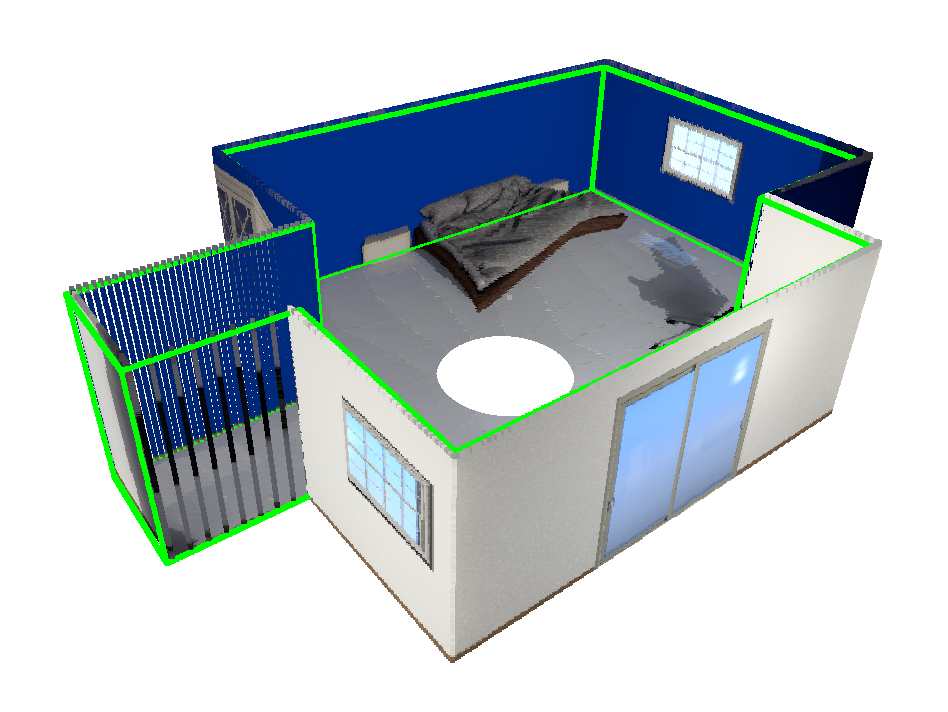}} \hfil
	\subfloat{\includegraphics[width=0.21\textwidth ,valign=c]{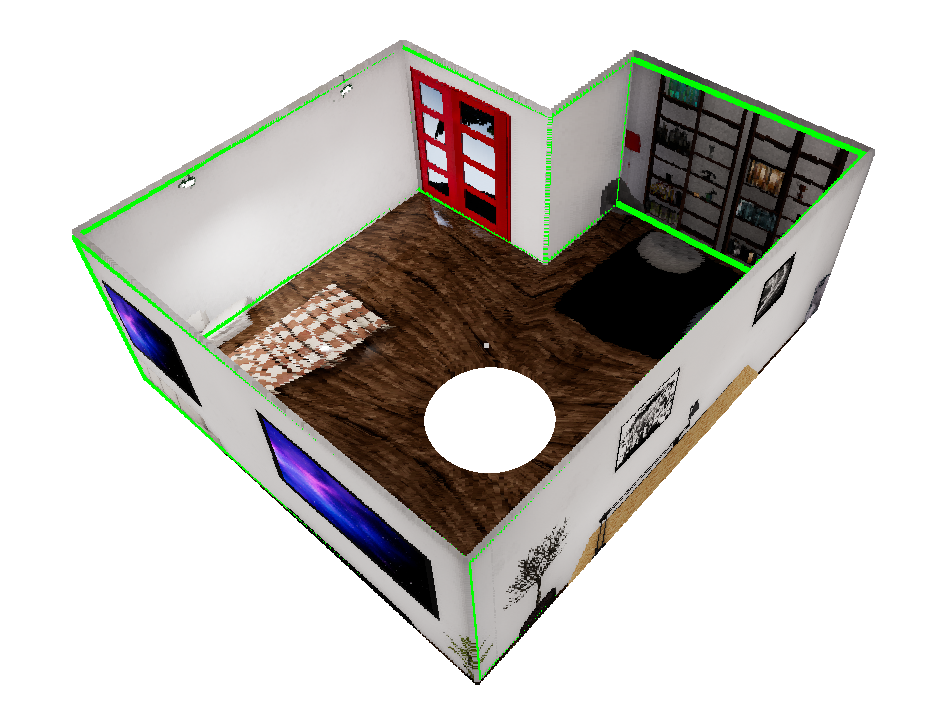}} \hfil
	\subfloat{\includegraphics[width=0.21\textwidth ,valign=c]{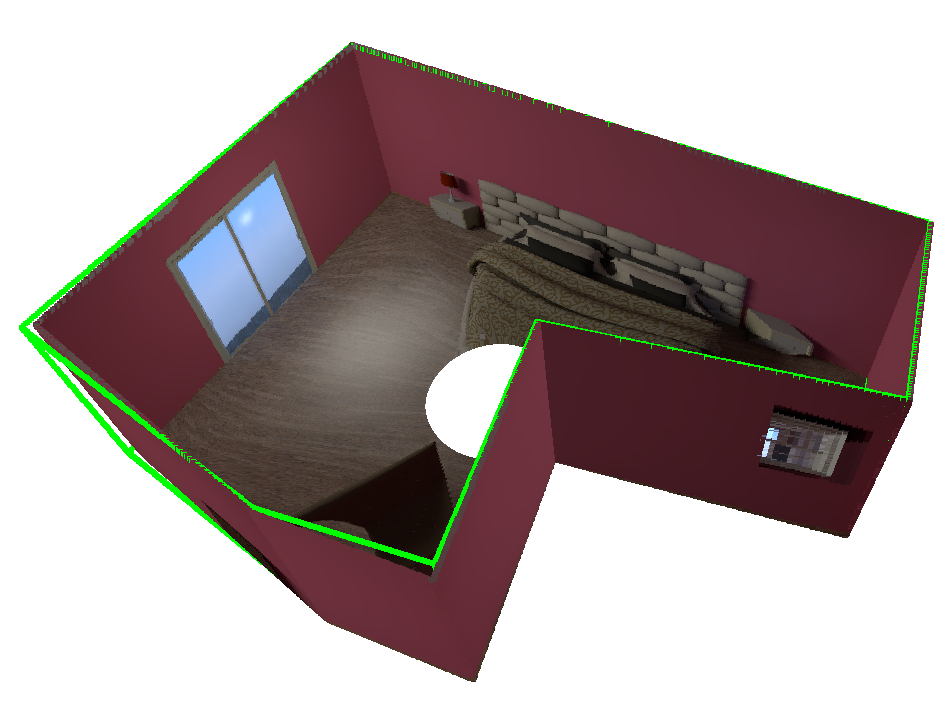}} \hfil
	\subfloat{\includegraphics[width=0.21\textwidth ,valign=c]{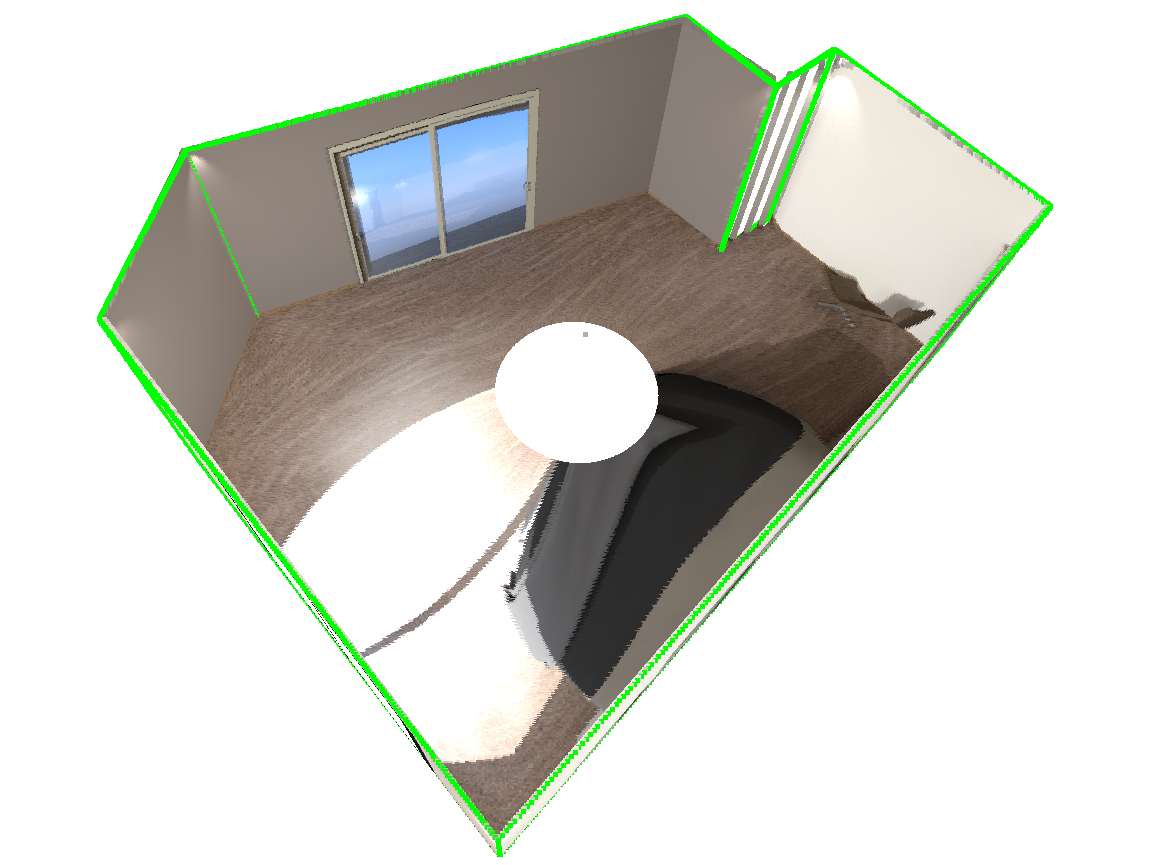}} 
	\caption{Examples of 3D reconstruction from the proposed method. We shown the non-central panorama and the 3D reconstruction. The green wire frame is the real 3D layout of the room.}
	\label{fig:results}
\end{figure*}

\begin{table}
\centering
\caption{Comparison of different methods of 3D layout recovery. }
\label{tab:comparison}
\begin{tabular}{l|cc|cc|}

\cline{2-5} 
 & \multicolumn{4}{|r|}{Manhattan World assumption} \\ \cline{2-5}
                   &   3D IoU  & \multirow{2}{*}{3D IoU}    &	\multirow{2}{*}{CEN}    &	\multirow{2}{*}{CE} \\ 
                   &    (u2s)  &                            &                           &                       \\ \hline
\multicolumn{1}{|l|}{CFL \cite{fernandez2020corners}}          
                        &	78.87	    & -     &	0.75	&	-	     \\
\multicolumn{1}{|l|}{HorizonNet \cite{sun2019horizonnet}}   
                        & 	82.66	    & -     &{\bf 0.69}	&	-	     \\
\multicolumn{1}{|l|}{AtlantaNet \cite{pintore2020atlantanet}}   
                        &	83.94	    & -     &	0.71	&	-	     \\
\multicolumn{1}{|l|}{\bf Ours }     
                        &   {\bf 93.88 }& {\bf 86.18} &	0.787   &  {\bf 0.223}     \\ 
\hline
 & \multicolumn{4}{|r|}{Atlanta World assumption}\\
\cline{1-5} 
\multicolumn{1}{|l|}{HorizonNet \cite{sun2019horizonnet}}   
                                    & 	73.53	    & -     &	-   	&	-	     \\
\multicolumn{1}{|l|}{AtlantaNet \cite{pintore2020atlantanet}}   
                                    &	80.01	    & -     &	-		&	-	     \\
\multicolumn{1}{|l|}{\bf Ours }     & {\bf 91.67}   & {\bf 76.17} &	{\bf 1.335}  &  {\bf 0.513}     \\ 
\hline
 \multicolumn{1}{c}{}& \multicolumn{2}{c}{\it higher is better} & \multicolumn{2}{c}{\it smaller is better} 
\end{tabular}
\end{table}

We have performed a set of experiments in order to evaluate our proposal and make a comparison with the state-of-the-art methods. The metrics used for the comparison are: 3D IoU, which refer to the 3D intersection over union of the predicted layout and the ground truth; 3D IoU(u2s), which refer to the up-to-scale intersection over union of the layout; CEN, which refer to the Corner Error Normalized computed as the L2 distance of the corners divided by the diagonal of the layout; CE, which refers to the Corner Error computed as the L2 distance of the corners in meters.

The comparison with state-of-the-art methods is not completely fair. The datasets used for the different methods are different, so the results can depend on the dataset used and not only on the method. Besides, our proposal only uses the image information in order to recover the 3D layout with the scale while the rest of the methods in the state of the art provide up-to-scale measures, relying on some measure in the environment for the 3D reconstruction, e.g. the camera height. Nevertheless, a summary of this comparison is shown in Table 1. 

These results show that our proposal outperforms the state of the art methods for Manhattan as well as Atlanta world assumptions. Besides, our method also recovers the scale of the layout without any prior assumption. Some examples of our results are shown in Fig. \ref{fig:results}, where different layouts are tested. We demonstrate that our method can handle quite challenging layouts, in different illumination conditions and world assumptions.

\section{Conclusions}

In this paper we have proposed two new solvers for indoor layout recovery from a single non-central circular panorama. We use a neural network for extracting the edges of structural lines from a non-central projection system and geometrically process the output in order to recover the 3D information of the layout. Our experiments show that our approach with non-central circular panoramas has better performance than state-of-the-art methods for Manhattan and Atlanta environments. In addition, our method can extract the scale of the room without any prior knowledge. 

\bibliographystyle{unsrtnat}
\bibliography{shbib}

\section*{ACKNOWLEDGMENT}

This work was supported by RTI2018-096903-B-100 (AEI/ FEDER, UE).

\end{document}